\documentclass[10pt,twocolumn,letterpaper]{article}

\usepackage[nocompress]{cite}
\usepackage{cvpr}
\usepackage{times}
\usepackage{epsfig}
\usepackage{graphicx}
\usepackage{amsmath}
\usepackage{amssymb}
\usepackage{subfigure}
\usepackage[font=small,labelfont=bf]{caption}
\usepackage[Algorithm]{algorithm}
\usepackage{algpseudocode}
\usepackage{setspace}
\usepackage{booktabs}
\floatname{algorithm}{Procedure}
\usepackage{color} 
\usepackage[labelsep=period]{caption}

\usepackage[pagebackref=true,breaklinks=true,letterpaper=true,colorlinks,bookmarks=false]{hyperref}

\cvprfinalcopy % *** Uncomment this line for the final submission

\def\cvprPaperID{2493} % *** Enter the CVPR Paper ID here

\ifcvprfinal\fi
\begin{document}

%%%%%%%%% TITLE
\title{Quality Aware Network for Set to Set Recognition}

\author{Yu Liu\\
SenseTime Group Limited\\
{\tt\small liuyuisanai@gmail.com}
\and
Junjie Yan\\
SenseTime Group Limited\\
{\tt\small yanjunjie@sensetime.com}
\and
Wanli Ouyang\\
University of Sydney\\
{\tt\small wanli.ouyang@gmail.com}
}

\maketitle
%\thispagestyle{empty}
%%%%%%%%% ABSTRACT
\begin{abstract}
This paper targets on the problem of set to set recognition, which learns the metric between two image sets.  Images in each set belong to the same identity.  Since images in a set can be complementary, they hopefully lead to higher accuracy in practical applications. However, the quality of each sample cannot be guaranteed, and samples with poor quality will hurt the metric. In this paper, the quality aware network (QAN) is proposed to confront this problem, where the quality of each sample can be automatically learned although such information is not explicitly provided in the training stage. The network has two branches, where the first branch extracts appearance feature embedding for each sample and the other branch predicts quality score for each sample. Features and quality scores of all samples in a set are then aggregated to generate the final feature embedding. We show that the two branches can be trained in an end-to-end manner given only the set-level identity annotation. Analysis on gradient spread of this mechanism indicates that the quality learned by the network is beneficial to set-to-set recognition and simplifies the distribution that the network needs to fit. Experiments on both face verification and person re-identification show advantages of the proposed QAN. The source code and network structure can be downloaded at GitHub\footnote{ \textcolor{blue}{\textit{https://github.com/sciencefans/Quality-Aware-Network}} Note that we are developing P-QAN (a fine-grained version of QAN, see Sec.\ref{diss}) in this repository. So the performance may be higher than that we report in this paper.}
\end{abstract}

%%%%%%%%% BODY TEXT
\section{Introduction}

\begin{figure}[!t]
  \centering
  \includegraphics[width=8cm]{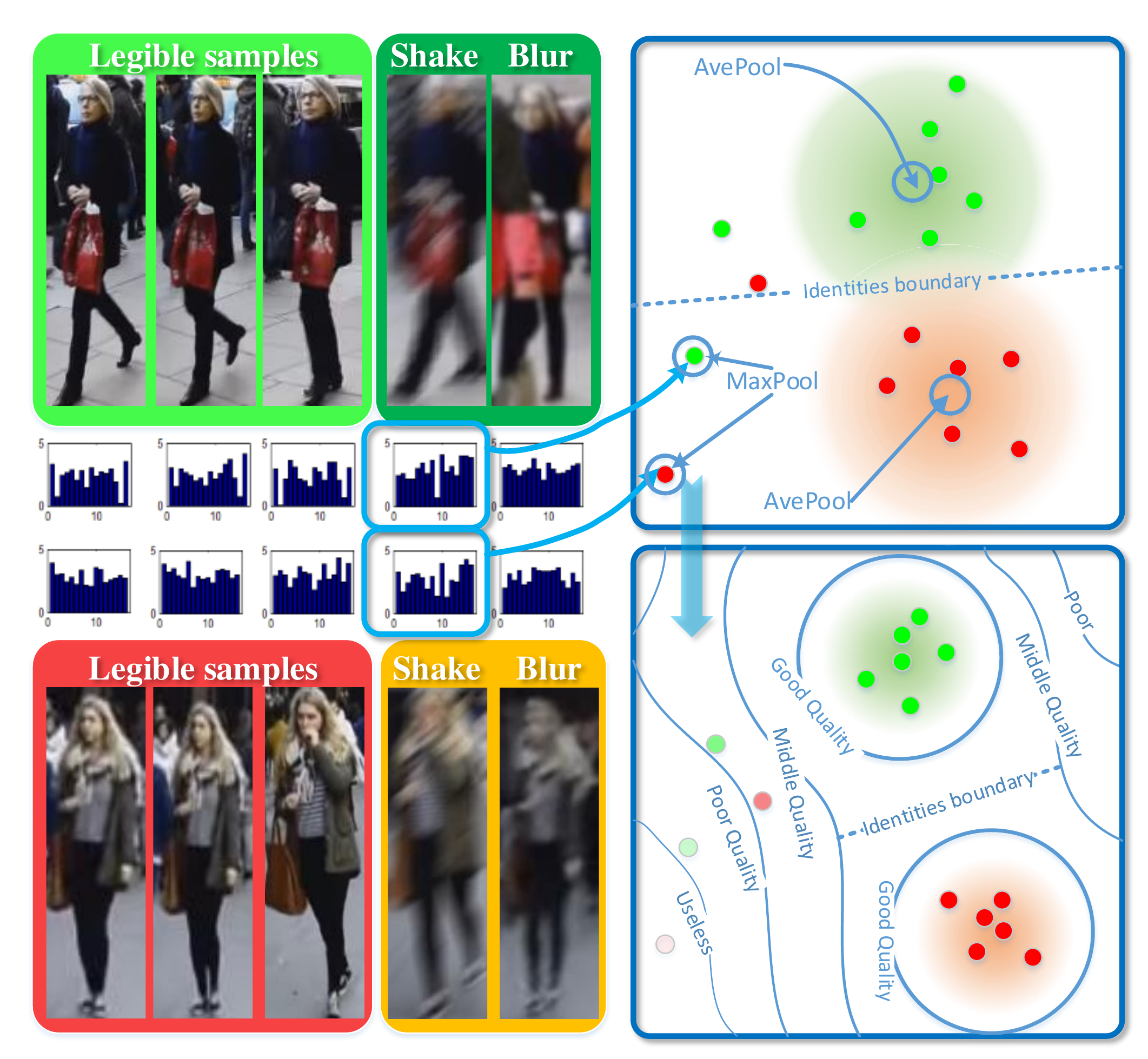}
  \caption{Illustration of our motivation, best viewed in color. \emph{\textbf{Left column:}} A classical puzzle in set-to-set recognition. Both set A (upper) and B (lower) contain noisy image samples caused by shake and blur. Their features (shown by histograms in middle row) are more similar to samples in other class than the inner class.  \emph{\textbf{Right column:}} Distributions and samples of two identities in hyperspace. Top: Due to the noisy, variances of two identities are large and they both have hard negative samples. Bottom: Quality aware network (QAN) weaken the noisy samples and narrow down identities' variances, which makes them more discriminative.}
  \label{fig:fig1}
\end{figure}

Face verification \cite{sun2014deep1,learnedlabeled,sun2014deep,taigman2014deepface,schroff2015facenet} and person re-identification \cite{gong2014person,farenzena2010person,zheng2011person,li2014deepreid} have been well studied and widely used in computer vision applications such as financial identity authentication and video surveillance. Both the two tasks need to measure the distance between two face or person images. Such tasks can be naturally formalized as a metric learning problem, where the distance of images from the same identity should be smaller than that from different identities. Built on large scale training data, convolutional neural networks and carefully designed optimization criterion, current methods can achieve promising performance on standard benchmarks, but may still fail due to appearance variations caused by large pose or illumination.

In practical applications, instead of one single image, a set of images for each identity can always be collected. For example, the image set of one identity can be sampled from the trajectory of the face or person in videos. Images in a set can be complementary to each other, so that they provide more information than a single image, such as images from different poses. The direct way to aggregate identity information from all images in a set can be simply max/average pooling appearance features of all images. However, one problem in this pooling is that some images in the set may be not suitable for recognition. As shown in Figure~\ref{fig:fig1}, both sets from left-top and left-bottom hold noisy images caused by shake or blur. If the noisy images are treated equally and max/average pooling is used to aggregate all images' features, the noisy images will mislead the final representation.

In this paper, in order to be robust to images with poor quality as described above and simultaneously use the rich information provided by the other images, our basic idea is that each image can have a quality score in aggregation. For that, we propose a quality aware network (QAN), which has two branches and then aggregated together. The first branch named \emph{feature generation part} extracts the feature embedding for each image, and the other branch named \emph{quality generation part} predicts quality score for each image. Features of images in the whole set are then aggregated by the final \emph{set pooling unit} according to their quality. 

A good property of our approach is that we do not supervise the  model by any explicit annotations of the quality. The network can automatically assign low quality scores to images with poor quality in order to keep the final feature embedding useful in set-to-set recognition. To implement that, an elaborate model is designed in which embedding branch and score generation branch can be jointly trained through optimization of the final embedding.  Specially in this paper, we use the joint triplet and softmax loss on top of image sets. The designed gradient of image set pooling unit ensures the correctness of this automatic process. 

Experiments indicate that the predicted quality score is correlated with the quality annotated by human, and the predicted quality score performs better than human in recognition. In this paper, we show the applications of the proposed method on both person re-identification and face verification. For person re-identification task, the proposed quality aware network improves top-1 matching rates over the baseline by 14.6\% on iLIDS-VID  and 9.0\% on PRID2011. For face verification, the proposed method reduces 15.6\% and 29.32\% miss ratio  when the false positive rate is 0.001 on YouTube Face and IJB-A benchmarks.

The main contributions of the paper are summarized as follows.
\begin{itemize}
\item The proposed quality aware network automatically generates quality scores for each image in a set and leads to better representation for set-to-set recognition.

\item We design an end-to-end training strategy and demonstrate that the quality generation part and feature generation part benefit from each other during back propagation.

%\item An adaptive cascade structure is designed to learn a better quality generator. It can adapt different classifiers to the images with different qualities.

\item  Quality learnt by QAN is better than quality estimated by human and we achieves new state-of-the-art performance on four benchmarks for person re-identification and face verification.
\end{itemize}

\section{Related work}

Our work is build upon recent advances in deep learning based person re-identification and unconstrained face recognition.  In person re-identification, \cite{li2014deepreid,xiao2016learning,zheng2016mars} use features generated by deep convolutional network and obtain state-of-the-art performance.  To learn face representations in unconstrained face recognition, Huang et al. \cite{Huang2012Learning} uses convolutional Restricted Boltzmann Machine while deep convolutional neural network is used  in \cite{taigman2014deepface, sun2014deep1}. Furthermore,  \cite{sun2015deeply,schroff2015facenet} use deeper convolutional network and achieved accuracy that even surpasses human performance. The accuracy achieved by deep learning on image-based face verification benchmark LFW\cite{learnedlabeled} has been promoted to 99.78\%. Although deep neural network has achieved such great performance on these two problems, in present world, unconstrained set-to-set recognition is more challenging and useful.

Looking backward, there are two different approaches handling set-to-set recognition. The first approach takes image set as a convex hull \cite{cevikalp2010face}, affine hull \cite{hu2011sparse} or subspace \cite{basri2011approximate, Huang2015Projection}. Under these settings, samples in a set distribute in a Hilbert space or Grassmann mainfold so that this issue can be formulated as a metric learning problem \cite{lu2015multi,yang2013face}.

Some other works degrade set-to-set recognition to point-to-point recognition through aggregating images in a set to a single representation in hyperspace. The most famous approach in this kind is the Bag of features \cite{lazebnik2006beyond}, which uses histogram to represent the whole set for feature aggregation. Another classical work is vector of locally aggregated descriptors (VLAD) \cite{jegou2010aggregating}, which aggregates all local descriptors from all samples. Temporal max/average pooling is used in \cite{wu2016deep} to integrate all frames' features generated by recurrent convolutional network. This method uses the 1st order statistics to aggregate the set. The 2nd order statistics is used in \cite{wang2012covariance, Zhu2013From} in assuming that samples follow Gaussian distribution. In \cite{hassner2016pooling}, original faces in a set are classified into 20 bins based on their pose and quality. Then faces in each bin are pooled to generate features and finally feature vectors in all bins are merged to be the final representation. \cite{yang2016neural} uses attention mechanism to summarize several sample points to a single aggregated point.

\label{sec:qualitynet}
\begin{figure*}[!htbp]
  \centering
  \includegraphics[width=17cm]{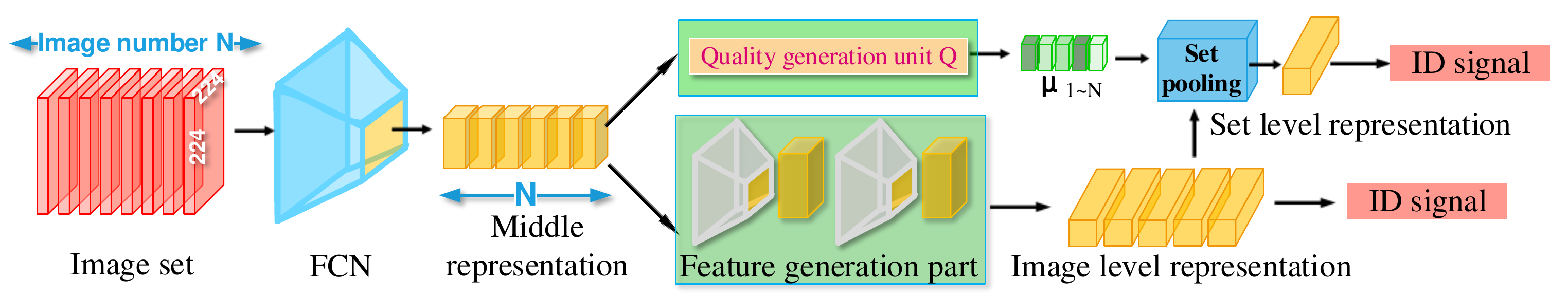}
  \caption{The end-to-end learning structure of quality aware net. The input of this structure is three image sets $S_{anchor}$, $S_{pos}$ and $S_{neg}$ belong to class $A$, $A$ and $B$. Each of them pass through the fully convolutional network (FCN) to generate the middle representations, which will be fed to quality generation part and feature generation part. The former generates quality score for each image and the latter generates final representation for each image. Then the scores and representations of all image will be aggregated by set pooling unit and the final representation of the image set will be produced. We use softmax-loss and triplet-loss to be the supervised ID signal. }
\label{figure_train}
\end{figure*}

The proposed QAN belongs to the second approach. It 
discards the dross and selects the essential information in all images. Different from recent works which learn aggregation based on fixed feature \cite{yang2016neural} or image\cite{hassner2016pooling}, the QAN learns feature representation and aggregation simultaneously.  \cite{goswami2014mdlface} proposed a similar quality aware module named ``memorability based frame selection'' which  takes ``visual entropy'' to be the score of a frame. But the score of a frame is defined by human and independent with feature generation unit. In QAN,  score is automatically learned and quality generation unit is joint trained with feature generation unit. Due to mutual benefit between the two parts during training, performance is improved significantly by jointly optimizing images aggregation parameter and images' feature generator.

\section{Quality aware network (QAN)}

In our work we focus on improving image set embedding model, which maps an image set $S=\{I_1, I_2, \cdots, I_N\}$ to an representation with fixed dimension so that image sets with different number of images are comparable with each other. Let $R_a(S)$ and $R_{I_i}$ denote representation of $S$ and $I_i$. $R_a(S)$ is determined by all elements in $S$, therefore it can be denoted as
\begin{equation}
\small
\label{3}
R_a(S) = \mathcal{F}(R_{I_1}, R_{I_2}, \cdots, R_{I_N}).
\end{equation}

The $R_{I_i}$ is produced by a feature extraction process, containing  traditional hand-craft feature extractors or convolutional neural network.  $\mathcal{F}(\cdot)$ is an aggregative function, which maps a variable-length input set to a representation of fixed dimension. The challenge is to find an optimized $\mathcal{F}(\cdot)$, which aggregate features from the whole image set to obtain the most discriminative representation. 
Based on notion that images with higher quality are easier for recognition while images with lower quality containing occlusion and large pose have less effect on set representation, we denote $\mathcal{F}(\cdot)$ as
\begin{equation}
\small
\label{5}
\mathcal{F}(R_{I_1}, R_{I_2}, \cdots, R_{I_N}) =\frac{\sum_{i=1}^N\mu_i R_{I_i}}{\sum_{i=1}^N\mu_i }
\end{equation}
\begin{equation}
\small
\label{6}
\mu_i = Q(I_i),
\end{equation}
where $Q(I_i)$ predicts a quality score $\mu_i$ for image $I_i$. So the representation of a set is a fusion of each images' features, weighted by their quality scores.

\subsection{QAN for image set embedding}
In this paper, feature generation and aggregation module is implemented through an end-to-end convolutional neural network named QAN as shown in Fig.~\ref{figure_train}. Two branches are splited from the middle of it. In the first branch, quality generation part followed by a set pooling unit composes the aggregation module. And in the second branch, feature generation part  generates images' representation. Now we introduce how an image set flows through QAN. At the beginning of the process, all images are sent into a fully convolutional network to generate middle representations. After that,
QAN is divided into two branches. The first one (upper) named quality generation part is a tiny convolution neural network (see Sec.~\ref{details_qgp} for details) which is employed to predict quality score $\mu$. The second one (lower), called feature generation part,  generates image representations $R_I$ for all images. $\mu$ and $R_I$ are aggregated at set pooling unit $\mathcal{F}$, and then pass through a fully connected layer to get the final representation $R_a(S)$. To sum up, this structure generates quality scores for images, uses these quality scores to weight images' representations and sums them up to produce the final set's representation.

\subsection{Training QAN without quality supervision}
\label{end2endtrain}
We train the QAN in an end-to-end manner. The data flow is shown in Fig.~\ref{figure_train}. QAN is supposed to generate discriminative representations for images and sets belonging to different identities. For image level training, a fully connection layer is established after feature generation part, which is supervised by Softmax loss $L_{class}$. For set level training, a set's representation $R_a(S)$ is supervised by $L_{veri}$ which is formulated as:
\begin{equation}
\small
\begin{aligned}
\label{7}
L_{veri} = &\left \lVert R_a(S_a)-R_a(S_p) \right \rVert^2 - &\left \lVert R_a(S_a)- R_a(S_n) \right \rVert^2 + \delta
\end{aligned}
\end{equation}

The loss function above is referred as \emph{Triplet Loss} in previous works \cite{schroff2015facenet}. We define $S_a$ as \emph{anchor set}, $S_p$ as \emph{positive set}, and $S_n$ as \emph{negative set}.  This function minimizes variances of intra-class samples while Softmax loss cannot guarantee that because softmax-loss directly optimizes the probability of each class, but not the discrimination of representation.

%\begin{equation}
%\footnotesize
%\label{difftrip}
%\begin{split}
%\frac{\partial L_{veri}}{\partial R_a(S_a)} = 2 \cdot (R_a(S_n)- R_a(S_p))  \\
%\frac{\partial L_{veri}}{\partial R_a(S_p)} = 2 \cdot (R_a(S_p)- R_a(S_a))  \\
%\frac{\partial L_{veri}}{\partial R_a(S_n)} = -2 \cdot (R_a(S_a)- R_a(S_n))
%\end{split}
%\end{equation}

% The gradient of Triplet loss tend to push the representations of different people further away while reducing the distance between representations of the same person.

Keeping this in mind, we consider the set pooling operation $\mathcal{F}$. The gradients back propagated through set pooling unit can be formulated as follows,
\begin{equation}
\small
\label{eq9}
\frac{\partial \mathcal{F}}{\partial R_{I_i}} =\frac{\partial R_a(S)}{\partial R_{I_i}} =\mu_i
\end{equation}
\begin{equation}
\small
\label{10}
\frac{\partial \mathcal{F}}{\partial \mu_i} =\frac{\partial R_a(S)}{\partial \mu_i} = R_{I_i} - R_a(S)
\end{equation}
So we can formulate propagation process of the final loss as
\begin{equation}
\small
\begin{aligned}
\label{11a}
\frac{\partial L_{veri}}{\partial R_{I_i}}
&= \frac{\partial R_a(S)}{\partial R_{I_i}} \cdot \frac{\partial L_{veri}}{\partial R_a(S)}  
&= \frac{\partial L_{veri}}{\partial R_a(S)} \cdot \mu_i
\end{aligned}
\end{equation}
\begin{equation}
\small
\begin{aligned}
\frac{\partial L_{veri}}{\partial \mu_i} &= \frac{\partial R_a(S)}{\partial \mu_i} \cdot (\frac{\partial L_{veri}}{\partial R_a(S)})^T  \\
&= \sum_{j=1}^D ( \frac{\partial L_{veri}}{\partial R_a(S)_j} \cdot ( x_{ij} - R_a(S)_j ) )
\label{11b}
\end{aligned}
\end{equation}
Where $D$ is the dimension of images' representation. We discuss how a quality score $\mu$ is automatically learned by this back propagation process.

\subsection{Mechanism for learning quality score}

\begin{figure}[!htbp]
  \centering
  \includegraphics[width=8cm]{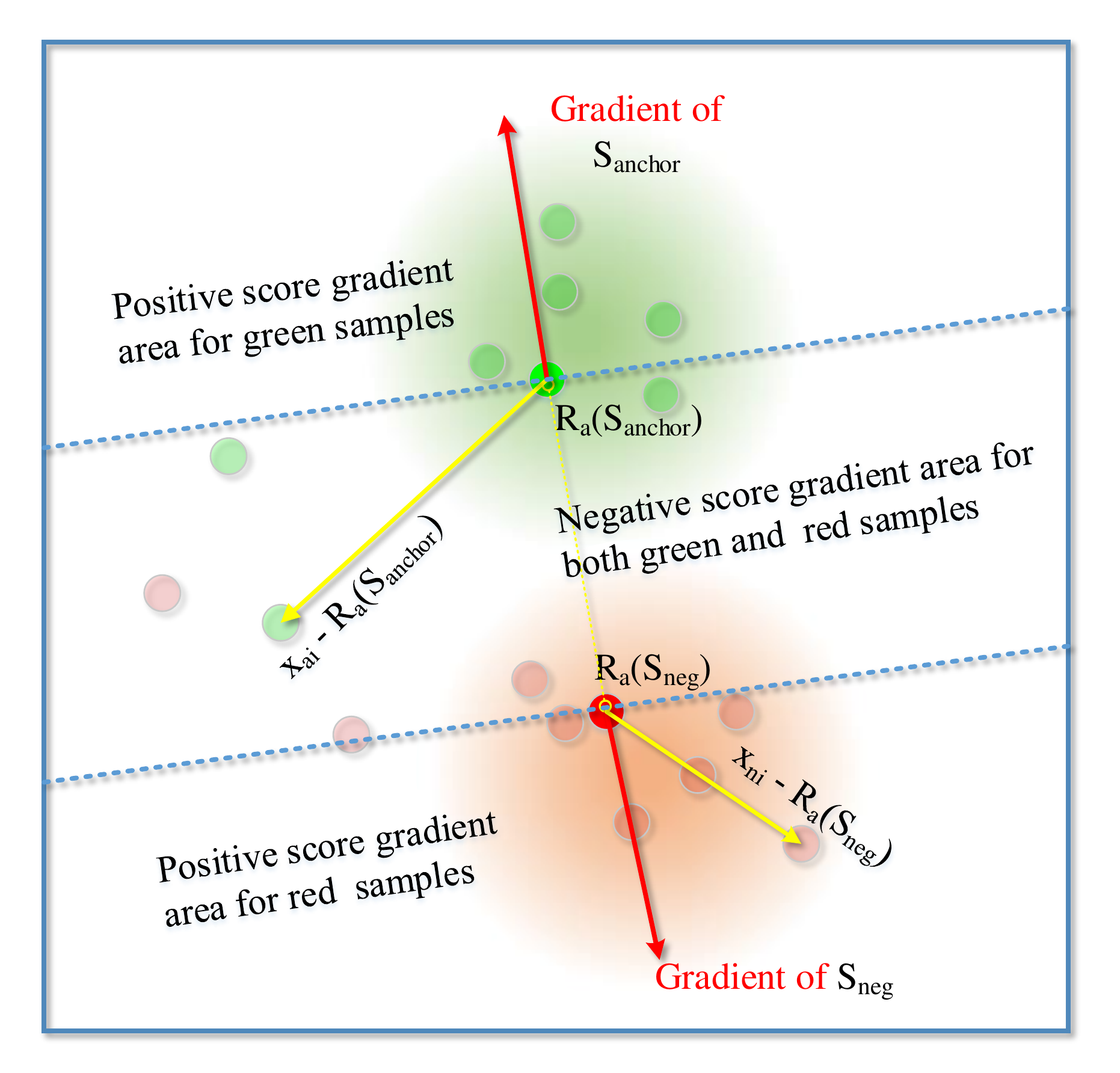}
  \caption{Two different identities in training, best viewed in color. Red translucent dots and green translucent dots indicate images in sets of two different identities. And the two solid dots denote the weighted centers of the two sets, which are also the representations of two sets $S_{anchor}$ and  $S_{neg}$. The gradients of $S_{anchor}$ and  $S_{neg}$ are shown with red arrows. The $x_{ni}$ and $x_{ai}$ are two image representations in two sets.} 
  \label{fig:gradient}
\end{figure}

\textbf{Automatic gradient of $\mathbf{\mu}$.}
After back-propagation through set pooling unit, gradient of $\mu_i$ with regard to $L_{veri}$ can be calculated according to the Eq.~\ref{11b}, which is the dot product of gradient from $R_a(S)$ and $R_{I_i}$. So if angle of $\nabla R_a(S)$ and $R_{I_i}$ belongs to ($-90^{\circ}$, $90^{\circ}$), $\mu_i$'s gradient will be positive. For example, as shown in Fig.~\ref{fig:gradient}, the angle of $\nabla R_a(S_{neg})$ and $x_{ni}-R_a(S_{neg})$ is less than $90^{\circ}$, so the $x_{ni}'s$ quality score $\mu_{ni}$ will become larger after this back propagation process. In contrast, the relative direction of $x_ai$ is in the opposite side of the gradient of $R_a(S_{anchor})$, making it obviously a hard sample, so its quality score $\mu_{ai}$ will tend to be smaller. Obviously, samples in the ``correct'' directions along with set gradient  always score higher in quality, while those in the ``wrong'' directions gain lower weight. For example in Fig.~\ref{fig:gradient}, green samples in the upper area and red samples in the lower area keep improving their quality consistently while in the middle area, sample's quality reduces. To this end, $\mu_i$ represents whether $i-th$ image is a good sample or a hard sample. This conclusion will be further demonstrated by experiments.

\textbf{$\mathbf{\mu}$ regulates the attention of $R_{I_i}$.}
The gradient of $R_{I_i}$ is shown in Eq.~\ref{11a} with a factor  $\mu_i$, together with the gradient propagated from Softmax loss. Since most of hard samples with lower $\mu_i$ are always poor images or even full of background noises, the factor $\mu_i$ in gradient of  $R_{I_i}$ weaken their harmful effect on the whole model. That is, their impact on parameters in feature generation part is negligible during back propagation. This mechanism helps feature generation part to focus on good samples and neglect ones, which benefits set-to-set recognition.

\subsection{Details of quality generation part}
\label{details_qgp}

\begin{figure}[!htbp]
  \centering
  \includegraphics[height=3.1cm]{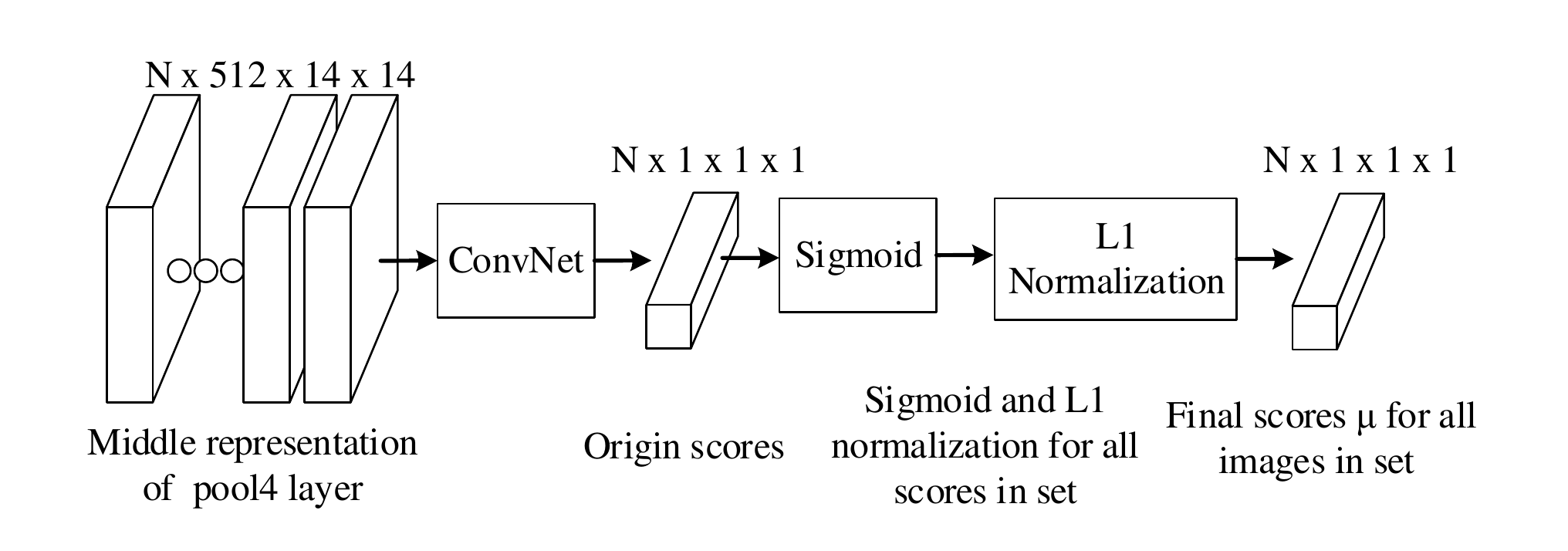}
  \caption{Structure of quality generation unit. The input of this unit is middle representations of a set which contains N images and it produces the normalized weights of all N images.}
  \label{qgu}
\label{figure2}
\end{figure}

\begin{figure*}[!ht]
  \centering
  \includegraphics[width=16cm]{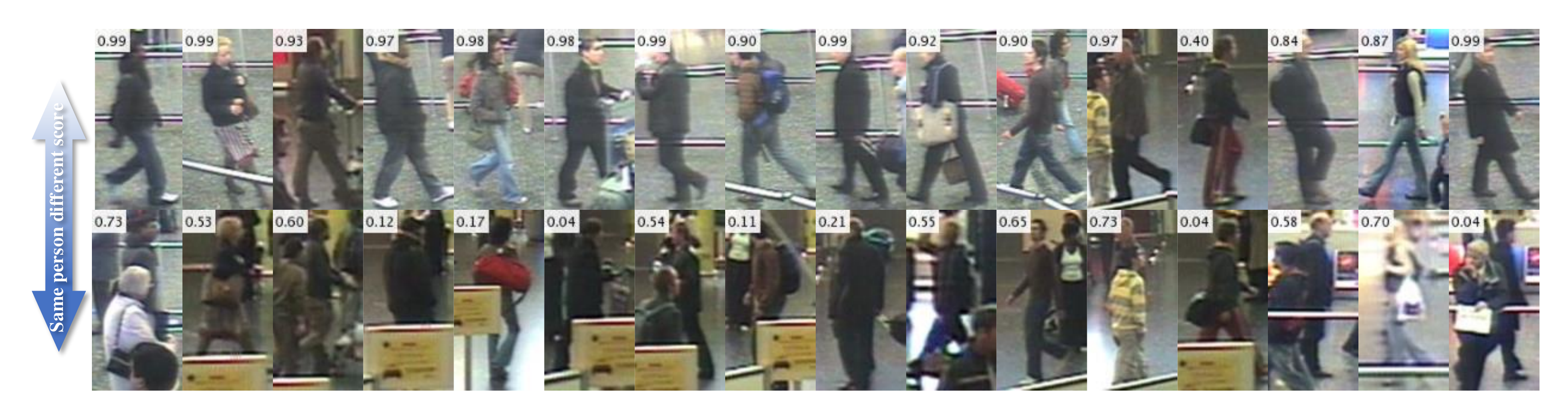}
  \includegraphics[width=16cm]{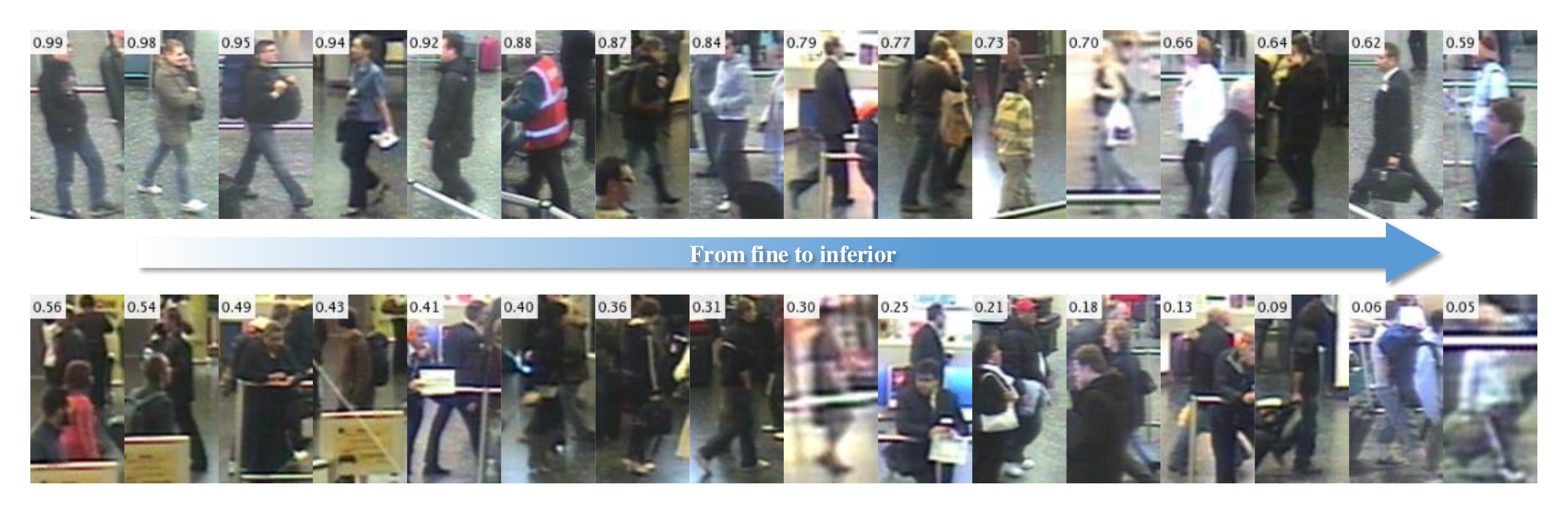}
  \caption{Samples with their qualities predicted by QAN, best viewed in color. \textbf{Top:} Comparison between two images from same person. From \textbf{up to down}, each column shows the two frames of a same person. The quality of the top one is better than the bottom one. \textbf{Bottom:} Random selected images in test set sorted by quality scores from \textbf{left to right}, best viewed in color.}
  \label{fig:samples}
\end{figure*}

In quality aware network (QAN), quality generation part is a  convolution neural network. We design different score generation parts start at different feature maps. We use QAN split at Pool4 as an instance. As shown in Fig.~\ref{qgu}, the output spatial of Pool4 layer is $512\times14\times14$. In order to generate a $1\times1$ quality score, the convolution part contains a 2-stride pooling layer and a final pooling layer with kernel size $7\times7$. A fully connected layer is followed by the final pooling layer to generate the original quality score. After that, the origin scores of all images in a set are sent to sigmoid layer and group L1-normalization layer to generate the final scores $\mu$. For QAN split at \texttt{Pool3}, we will add a block containing three 1-stride convolution layer and a 2-stride pooling layer at the beginning of quality generation unit.

\section{Experiments}
\label{others}
In this section, we first explore the meaning of the quality score learned by QAN. Then QAN's sensitivity to level of feature is analysed. Based on above knowledge, we evaluate QAN on two human re-identification benchmarks and two unconstrained face verification benchmarks. Finally, we analyse the concept learned by QAN and compare it with score labelled by human.
%\begin{figure}[h]
%  \label{fig-ped}
%  \centering
%  \includegraphics[width=12cm]{c5_experiment/dataset.pdf}
%  \caption{Samples of PRID2011 dataset and iLIDS-VID %dataset. Images in first row and second row belongs to %same person. So as images in third and fourth row.}
%\end{figure}

\subsection{What is learned in QAN?}

% Todo: fix this figure with typical samples
\textbf{Qualitative analysis} 
We visualize images with their $\mu$ generated by QAN to explore the meaning of $\mu$. Instances of same person with different qualities are shown in the first two rows in Fig.~\ref{fig:samples}. All images are selected from test set. The two images in the same column belong to a same person. The upper images are random selected from images with quality scores higher than 0.8 and the lower images are selected from images with quality scores lower than the corresponding higher one. It is easy to find that images with deformity, superposition, blur or extreme light condition tend to obtain lower quality scores than normal images.

The last two rows in Fig.~\ref{fig:samples} give some examples of other images random selected from test set. They are sorted by their quality scores from left to right. We can observe that instances with quality scores larger than 0.70 are easy to recognize by human while the others are hard. Especially many of hard images include two or more bodies in the center and we can hardly discriminate which one is the right target. 

\textbf{Quantitative analysis} 
In order to measure the relationship between the quality labelled by human and $\mu$ predicted by QAN, 1000 images in YouTube Face are selected randomly and the quality of them are rated subjectively by 6 volunteers, where each volunteer estimates a quality score for each image, ranging from 0 to 1. All the ratings of each volunteer are aligned by logistic regression. Then the 6 aligned scores of each image are averaged and finally normalized to $[0,1]$ to get the final quality score from human. %For each identity, Fig.~\ref{fig:ratings} shows five identities' faces in the whole set. We can see the score distributions of human and QAN are correlated. Faces with occlusion, blur, miss-align, abnormal pose and expression have lower scores both in human's score and QANs' score.

\begin{figure}[!htb]
\minipage{0.5\textwidth}
\centering
  \includegraphics[width=9cm]{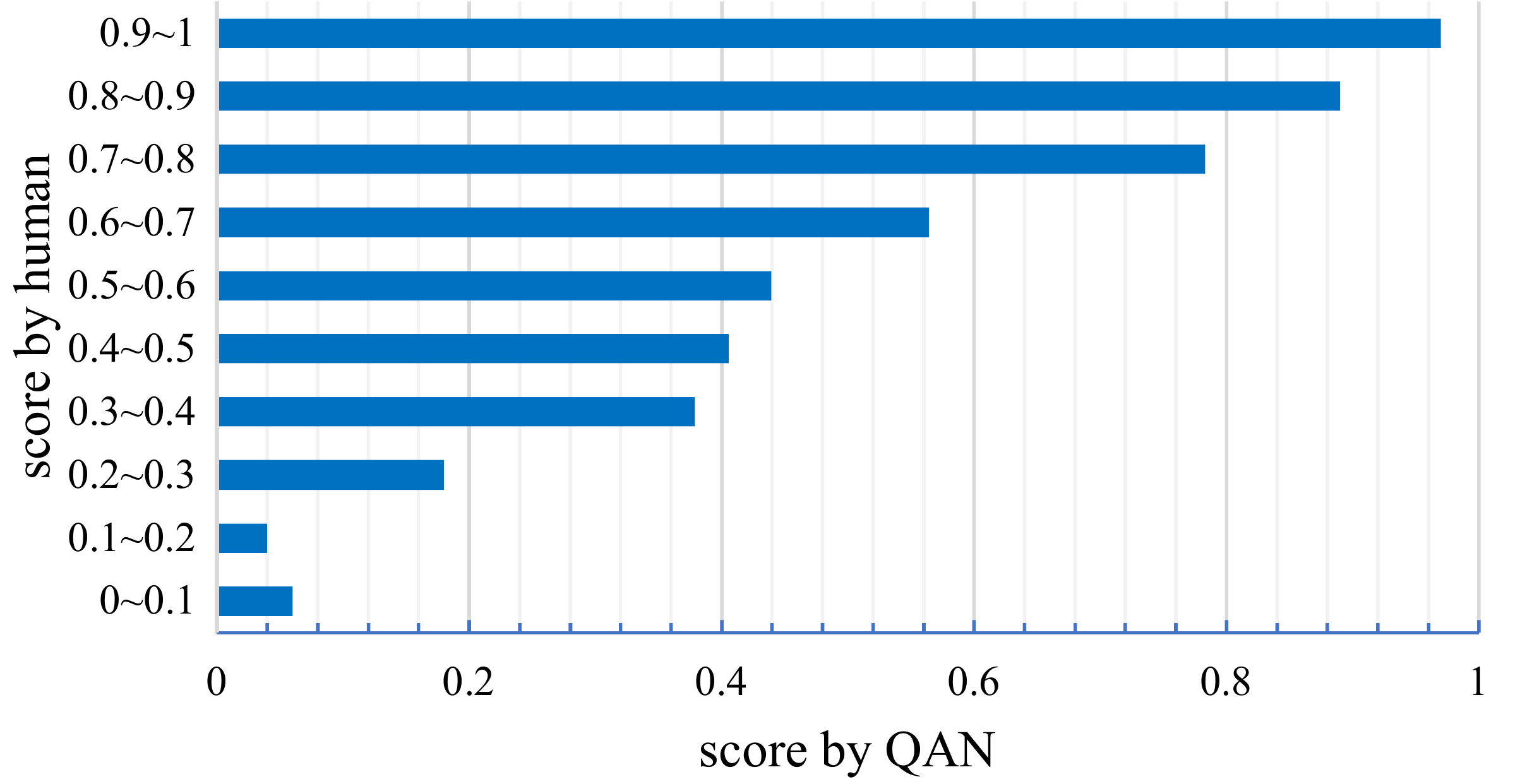}
  \caption{Comparison of qualities estimated by human and predicted by QAN.}
  \label{fig:quality}
\endminipage\hfill

%\minipage{0.32\textwidth}
%  \includegraphics[width=5cm]{c5_experiment/distribute1.pdf}
%  \caption{Distance distribution of features drawn from the output of QAN.}
%  \label{fig:subfig:a}
%\endminipage\hfill
%\minipage{0.32\textwidth}%
%  \includegraphics[width=5cm]{c5_experiment/distribute2.pdf}
%   \caption{Distance distribution of features directly drawn from feature generation module.}
%  \label{fig:subfig:b}
%\endminipage
\end{figure}

We divide the images into ten partitions based on human's score as shown in Fig.~\ref{fig:quality}. In which we show the corresponding quality statistics generated by QAN. It is obvious that the scores given by the QAN are strongly correlated with human-defined quality. We further analyse the 499,500 image pairs from these 1000 images and ask human and QAN to select the better one in each pair.  Result shows that the decision made by QAN has 78.1\% in common with human decision.

\subsection{Person re-identification}
\textbf{Datasets.} For person re-identification, we collect 134,942 frames with 16,133 people and 212,726 bounding boxes as the training data. Experiments are conducted on PRID2011~\cite{hirzer11a} and iLiDS-VID\cite{wang2014person} datasets. PRID2011 contains frames in two views captured at different positions of a street. \texttt{CameraA} has 385 identities while \texttt{CameraB} has 749 identities, and the two videos have a overlap of 200 people. Each person has 5 to 675 images, and the average number is 100. iLIDS-VID dataset has 300 people, and each person has two sets also captured from different positions. Each person has 23 to 192 images.

\textbf{Evaluation procedure.}
The results are reported in terms of \emph{Cumulative Matching Characteristics} (CMC) table, each column in which represents matching rate in a certain top-N matching.
Two settings are used for comprehensive evaluation. In the first setting, we follow the state-of-the-art method described in \cite{you2016top} and \cite{wang2016person}. The sets whose frame number is larger than 21 are used in PRID2011, and all the sets in iLIDS-VID are used. Each dataset is divided into two parts for fine-tuning and testing, respectively. For the testing set, sets form \texttt{CameraA} are taken as probe set while sets from \texttt{CameraB} are taken as the gallery. The final number is reported as the average of ``10-fold cross validation''. In the second setting, we conduct cross-dataset  testing. Different from the first setting, we ignore the finetuning process and use all data to test our model. That is, in PRID2011, the first 200 people from \texttt{CameraA} serve as probes, and all sets from \texttt{CameraB} are used as the gallery set. In iLIDS-VID,  \texttt{CameraA} are used as the probe set, and Camera B serve as gallery set.

\textbf{Baseline.}
We implement two baseline approaches. In the first baseline, we use average pooling to aggregate all images' representations. In the second baseline, a minimal cosine distance between two closures is used to be their similarity.

\subsubsection{Evaluation on common setting}

Results of evaluation obeying ``10-fold cross validation'' on  PRID2011 and iLIDS-VID are shown in Table~\ref{tab1} and Table~\ref{tab2}. Benefiting from the large scale training dataset, our CNN+AvePool and CNN+Min(cos) baselines are close to or even better than the state-of-the-art. Notice that most of the leading methods listed in table consider both appearance and spatio-temporal information while our method only considers appearance information. On PRID2011 dataset, QAN increase  top-1 matching rate by 11.1\% and 29.4\% compared with CNN+AvePool and CNN+Min(cos). On iLIDS-VID dataset, inherent noise is much more than that in PRID2011, which significantly influence the accuracy of CNN+Min(cos) since operator ``Min(cos)'' is more sensitive than ``AvePool'' to noisy samples . However, QAN achieves more gain on this noisy dataset. It increase top-1 matching rate by 12.21\% and 37.9\%. 

\begin{table}[!htb]
\normalsize
%\begin{scriptsize}
  \centering
  \begin{tabular}{l|c|c|c|c}
\hline
      \multicolumn{5}{c}{PRID2011}\\
    %\cmidrule{1-2}
\hline
       Methods & CMC1 &CMC5&CMC10 & CMC20 \\
\hline
       QAN 	& \textbf{90.3} & \textbf{98.2} & \textbf{99.32} & \textbf{100.0}  \\
       CNN+AvePool 		& 81.3 & 96.6 & 98.5 & 99.6 \\
       CNN+Min(cos) 		& 69.8 & 91.3 & 97.1 & 99.8  \\
\hline
       CNN+RNN\cite{wu2016deep} & 70 & 90 & 95 & 97  \\
       STFV3D\cite{liu2015spatio} & 42.1 & 71.9 & 84.4 & 91.6 \\
       TDL\cite{you2016top} 				& 56.7 & 80.0 & 87.6 & 93.6 \\
       eSDC\cite{wang2016person} 	& 48.3 & 74.9 & 87.3 & 94.4  \\
       DVR\cite{wang2016person} 	& 40.0 & 71.7 & 84.5 & 92.2 \\
       LFDA\cite{pedagadi2013local}	& 43.7 & 72.8 & 81.7 & 90.9  \\
       KISSME\cite{koestinger2012large}	& 34.4 & 61.7 & 72.1 & 81.0  \\
       LADF\cite{li2013learning} 				& 47.3 & 75.5 & 82.7 & 91.1 \\
       TopRank\cite{li2014top} 			& 31.7 & 62.2 & 75.3 & 89.4  \\
\hline
  \end{tabular}
  \caption{Comparison of QAN, \texttt{AvePool}, \texttt{Min(cos)} and other state-of-the-art methods on PRID2011, where the number represents the cumulative matching rate in CMC curve.}
%\end{scriptsize}
  \label{tab1}
\end{table}

\begin{table}[!htb]
\normalsize
%\begin{scriptsize}
  \centering
  \begin{tabular}{l|c|c|c|c}
\hline
     \multicolumn{5}{c}{iLIDS-VID}\\
    %\cmidrule{1-2}
\hline
       Methods & CMC1 &CMC5&CMC10 & CMC20   \\
\hline 

       QAN 	& \textbf{68.0} & \textbf{86.8} & 95.4 & 97.4  \\
       CNN+AvePool 		& 60.6 & 84.9 & 89.8 & 93.6 \\
       CNN+Min(cos) 		& 49.3 & 79.4 & 88.2 & 91.9  \\
\hline
       CNN+RNN\cite{wu2016deep} & 58 & 84 & 91 & 96 \\
       STFV3D\cite{liu2015spatio} & 37.0 & 64.3 & 77.0 & 86.9  \\
       TDL\cite{you2016top} 				& 56.3 & 87.6 & \textbf{95.6} & \textbf{98.3} \\
       eSDC\cite{wang2016person} 	& 41.3 & 63.5 & 72.7 & 83.1  \\
       DVR\cite{wang2016person} 				& 39.5 & 61.1 & 71.7 & 81.0 \\
       LFDA\cite{pedagadi2013local} 				& 32.9 & 68.5 & 82.2 & 92.6 \\
       KISSME\cite{koestinger2012large} 				& 36.5 & 67.8 & 78.8 & 87.1  \\
       LADF\cite{li2013learning} 				& 39.0 & 76.8 & 89.0 & 96.8  \\
       TopRank\cite{li2014top} 			& 22.5 & 56.1 & 72.7 & 85.9 \\
\hline
  \end{tabular}
  \caption{Comparison of QAN, \texttt{AvePool}, \texttt{Min(cos)} and other human re-identification methods on iLIDS-VID, where the number represents the cumulative matching rate on CMC curve.}
%\end{scriptsize}
  \label{tab2}
\end{table}

Based on these two experiments, QAN significantly outperforms two baselines on both datasets. It also performs better than many state-of-the-art approaches and pushes top-1 matching rate 20.3\% higher than previous best CNN+RNN\cite{wu2016deep} on PRID2011 and 10\%  on iLIDS-VID. The performance gain is more significant on noisy iLIDS-VID dataset, which meets the expectation and proves QAN's ability to deal with images of poor quality.

\begin{table}[ht]
\normalsize
  \centering
  \begin{tabular}{l|c|c|c|c}
    \hline
      \multicolumn{5}{c}{PRID2011}\\
    \hline
       Methods & CMC1 &CMC5&CMC10 & CMC20  \\
    \hline
       QAN 		  & \bf{34.0} & \bf{61.3} & \bf{74.0} & \bf{83.1}  \\
       CNN+AvePool 		& 29.4 & 57.5 & 68.8 & 80.2  \\
       CNN+Min(L2) 		& 28.5 & 57.1 & 67.1 & 78.6  \\
    \hline
    CNN+RNN\cite{wu2016deep} & 28 & 57 & 69 & 81 \\
    \hline
  \end{tabular}
  \caption{Cross-dataset performance of QAN on PRID2011, where the number represents the cumulative accuracy on CMC curve.}
  \label{tab_cross_prid}
\end{table}

\begin{table}[ht]
\normalsize
  \centering
  \begin{tabular}{l|c|c|c|c}
    \hline
      \multicolumn{5}{c}{iLIDS-VID}\\
    \hline
       Methods & CMC1 &CMC5&CMC10 & CMC20  \\
    \hline
       QAN		& \bf{47.7} & \bf{70.4} & \bf{83.9} & \bf{91.3} \\
       CNN+AvePool 		& 44.1 & 65.8 & 78.5 & 88.9 \\
       CNN+Min(L2) 		& 41.9 & 61.7 & 75.5 & 79.5  \\
    \hline
  \end{tabular}
  \caption{Cross-dataset performance of QAN on iLIDS-VID, where the number represents the cumulative accuracy on CMC curve.}
  \label{tab_cross_ilids}
\end{table}

\subsubsection{Dataset cross evaluation} To prevent our model from over-fitting the quality distribution of test set, we conduct dataset cross evaluation. We extract set representation of iLIDS-VID and PRID2011 directly using trained QAN without fine-tuning. The QAN representation is then evaluated for CMC scores. Table \ref{tab_cross_prid} and \ref{tab_cross_ilids} shows the results of QAN and the two baselines. It can be found that the QAN is robust even in cross-dataset setting. It improves top-1 matching by 15.6\% and 8.2\% compared to the baselines. This result shows that the quality distribution learned from different datasets by QAN is able to generalize to other datasets.

\begin{figure*}[!htb]
\minipage{0.32\textwidth}
\center
  \includegraphics[height=4cm]{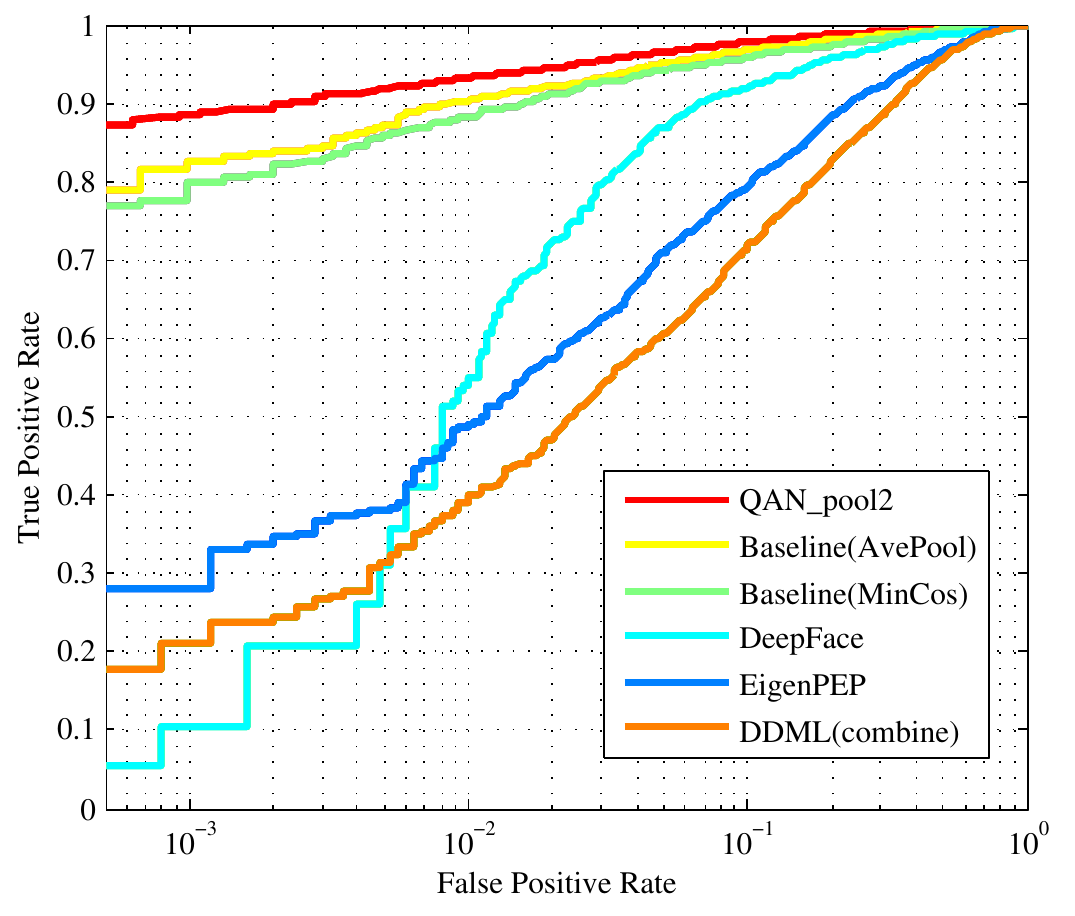}
  \caption{Average ROC curves of different methods on YouTube Face Dataset}
  \label{fig:rocs}
\endminipage\hfill
%\minipage{0.32\textwidth}
%\center
%  \includegraphics[width=5.5cm]{c5_experiment/cascade_num-tpr.pdf}
%  \caption{QAN's sensitivity to cascade number, analysed on YouTube Face Dataset.}
%  \label{fig:casnum2tpr}
%\endminipage\hfill
\minipage{0.35\textwidth}%
\center
  \includegraphics[height=4cm]{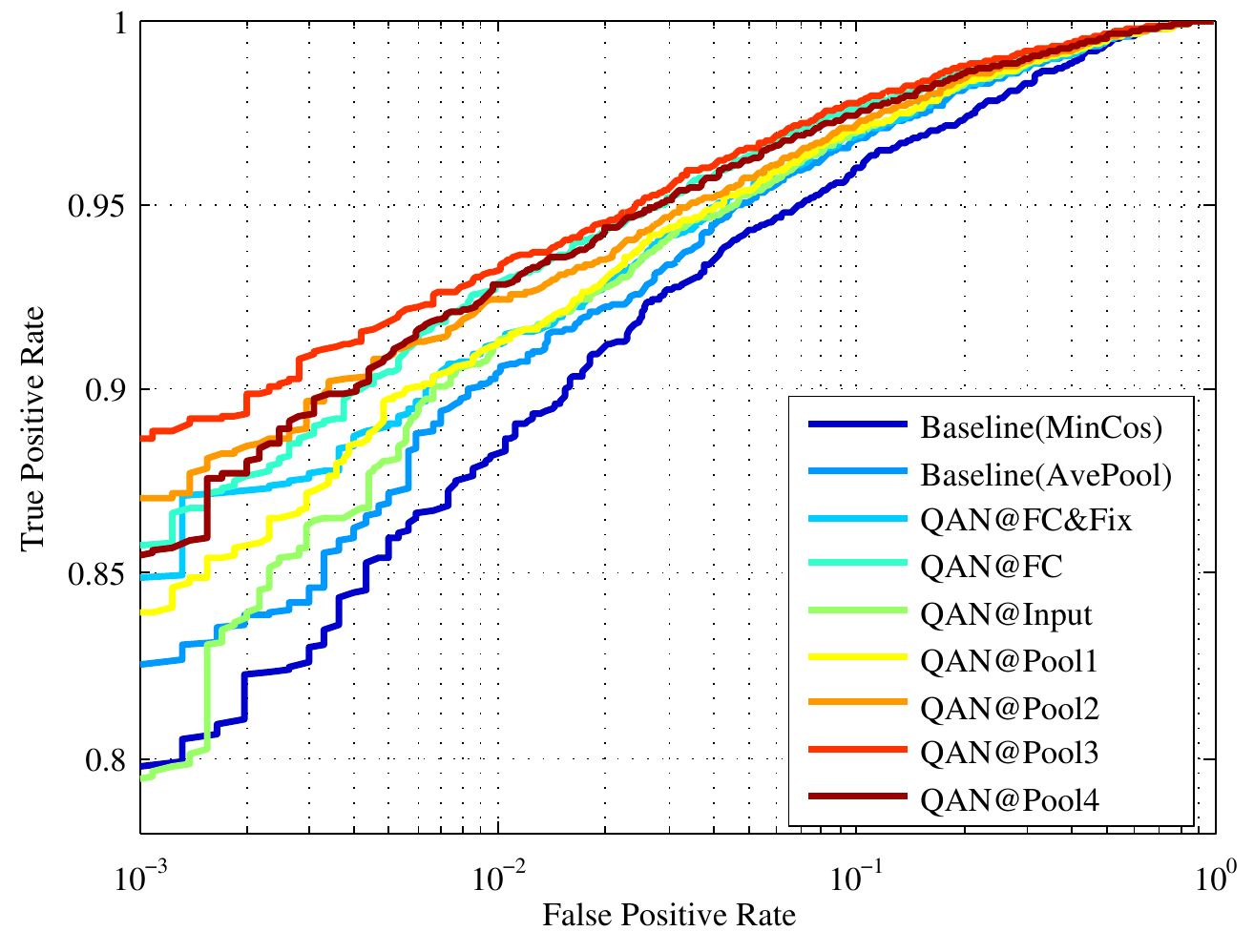}
  \caption{ROC results for score generation part learned by different level of feature.}
  \label{fig:qanloc}
\endminipage
\minipage{0.32\textwidth}
\centering
  \includegraphics[height=4cm]{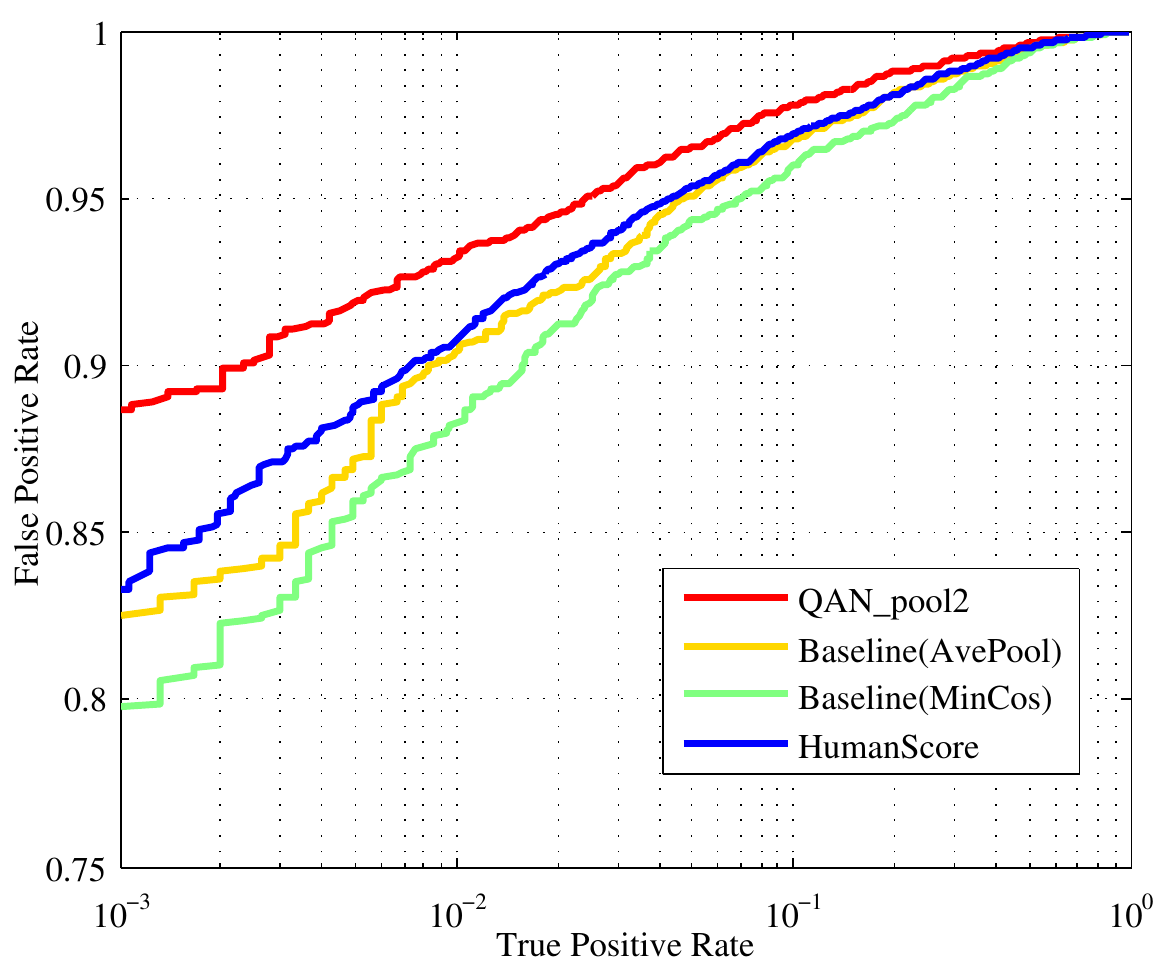}
  \caption{QAN with human score performs better than the two baselines but worse than that scored by network.\label{fig:roc4321}}
\endminipage

\end{figure*}

\subsection{Unconstrained face verification}
\textbf{Datasets.} For face verification, we train our base model on extended version of VGG Face dataset~\cite{Parkhi15}, in which we extend the identity number from 2.6K to 90K and image number from 2.6M to 5M. The model is evaluated on YouTube Face Database~\cite{wolf2011face} and IARPA Janus Benchmark A (IJB-A) dataset. YouTube Face contains 3425 videos of 1595 identities. It is challenging in that most faces are blurred or has low resolution. IJB-A dataset contains 2042 videos of 500 people. Faces in IJB-A have large pose variance.

\textbf{Evaluation procedure.} We follow the 1:1 protocol in both two benchmarks and evaluate results using receiver operating characteristic (ROC) curves. Area under curve (AUC) and accuracy are two important indicators of the ROC. The datasets are evaluated using 10-fold cross-validation.

\textbf{Training details.}
All faces in training and testing sets are detected and aligned by a multi-task region proposal network as described in \cite{chen2016supervised}. Then we crop the face regions and resize them to $256\times 224$. After that, a convolutional neural networks with $256\times 224$ inputs are used for face verification. It begins with a 2-stride convolution layer, followed by 4 basic blocks, while each block has three 1-stride convolution layers and one 2-stride pooling layers. After that, a fully connected layer is used to get the final feature. Quality generation branch is built on top of the third pooling layer, where the spatial size of middle representation response is $256\times 16 \times 14$. We pre-train the network supervised by classification signal and then train the whole QAN.

\subsubsection{Results on YouTube Face and IJB-A benchmark}

\begin{table}[h]
\normalsize
  \centering
  \begin{tabular}{l|c|c}
    \hline
    Method & Accuracy(\%) &AUC \\
    \hline
       QAN   & \bf{96.17$\pm$ 0.09\%} & \bf{99.14$\pm$ 0.12\%} \\
       CNN+AvePool 	 & 95.46$\pm$ 0.07\% & 98.66$\pm$ 0.04\% \\
       CNN+Min(cos)  & 94.87$\pm$ 0.10\% & 98.37$\pm$ 0.06\% \\
    \hline
       NAN\cite{yang2016neural}			 & 95.52$\pm$0.06\% & 98.7\% \\
       FaceNet\cite{schroff2015facenet} 	& 95.12$\pm$0.39\% & - \\
       DeepID2+\cite{sun2015deeply}			& 93.2$\pm$0.2\% & - \\
       DeepFace-single\cite{taigman2014deepface}		& 91.4$\pm$1.1\% & 96.3\% \\
       EigenPEP\cite{li2014eigen}			& 84.8$\pm$1.4\% & 92.6\% \\
    \hline
  \end{tabular}

  \caption{Average accuracy and AUC of QAN on YouTube Face dataset, compared with baselines and other state-of-the-arts.}
  \label{tab:ytf}
\end{table}

\begin{table}[!htbp]
\footnotesize
  \centering
  \begin{tabular}{l|c|c|c}
    \hline
       TPR@FPR & 1e-3 & 1e-2  & 1e-1 \\
    \hline
       QAN 	& \bf{89.31$\pm$3.92\%} & \bf{94.20$\pm$1.53\%} & \bf{98.02$\pm$0.55\%} \\
       CNN+AvePool 		& 85.30$\pm$3.48\% & 93.81$\pm$1.44 & 97.85$\pm$0.61\% \\
       CNN+Min(cos) 	& 82.74$\pm$3.61\% & 92.06$\pm$1.98 & 97.29$\pm$0.67\% \\
    \hline
       NAN\cite{yang2016neural}		&78.5$\pm$2.8\% &89.7$\pm$1.0\% &95.9$\pm$0.5\% \\
       DCNN+metric\cite{chen2015end} 	& - & 78.7$\pm$4.3\% & 94.7$\pm$1.1\% \\
       LSFS\cite{wang2015face}				& 51.4$\pm$6.0\% & 73.3$\pm$3.4\% & 89.5$\pm$1.3\% \\
       OpenBR\cite{klontz2013open}				& 10.4$\pm$1.4\% & 23.6$\pm$0.9\% & 43.3$\pm$0.6\% \\
    \hline
  \end{tabular}
  \caption{TPRs of QAN at specific FPRs on IJB-A dataset, compared with baselines and other state-of-the-arts.}
  \label{tab:ijb}
%\end{scriptsize}
\end{table}

On YouTube Face dataset, it can be observed in Fig.~\ref{fig:rocs} and Table~\ref{tab:ytf} that the accuracy and AUC of our baselines are similar with the state-of-the-art methods such as FaceNet and NAN. Based on this baseline, QAN further reduces 15.6\% error ratio. Under ROC evaluation metric, QAN surpasses NAN by 8\% and DeepFace by 80\% at 0.001 FPR (false positive rate), which ensembles 25 models.

On IJB-A dataset, QAN significantly outperforms the state-of-the-art algorithm NAN by 10.81\% at 0.001 FPR, 4.5\% at 0.01 FPR and 2.12\% at FPR=0.1, as shown in Table~\ref{tab:ijb}. Compared with average pooling baseline, QAN reduces false negative rate at above three FPRs by 29.32\%, 6.45\% and 7.91\%.

Our experiments on the two tasks show that QAN is robust for set-to-set recognition. Especially on the point of low FPR, QAN can recall more matched samples with less errors.

\subsection{ Quality by QAN VS. quality by human}
\label{sec:humanscorebetter}
There is no explicit supervision signals for the cascade score generation unit in training. So another problem arises: is it better to use human-defined scores instead of letting the network learn itself? In YouTube Face experiment, we replace the quality score $Q(I)$ with volunteer-rated score and get the following result in Fig.~\ref{fig:roc4321}, which is better than the two baselines but inferior to the result of original QAN. It shows that $Q$ is similar with human thoughts, but more suitable for recognition. Quality score by human can also enhance the accuracy but is still worse than QAN's.

\subsection{Diagnosis experiments}
%Cascade number in quality generation part and 
Level of middle representation may affect the performance of QAN. We use YouTube Face to analyse this factor by comparing different configurations.

%\textbf{Cascade number.} We examine ROC performance with respect to the cascade number.  Fig.~\ref{fig:casnum2tpr} shows that the TPRs at FPR=1E-1 and FPR=1E-2 converge at 2-cascade while TPR at FPR=1E-3 converges at 3-cascade. This indicates that quality generation parts at higher levels can learn better weight metrics than that at low levels. Finally, the performance decreases at the 4-th cascade due to the over-fitting.

%\textbf{Feature level.} To discover which level of feature is better for weight learning in QAN, we compare 3-cascade QANs with different configurations. 
In the first configuration, the weight generation part is connected to the image. In the second to fifth configurations, weight generation part is set after four pooling layers in each block, respectively. In the sixth configuration, we connect weight generation part to a fully connected layer. For the final configuration, we fix all parameters before the final fully connection layer in the sixth configuration and only update parameters in weight generation part, which is taken as the seventh structure. To minimize the influence by parameters' number, the total size of different models is restricted to the same by changing the channel number. 

Results are shown in Fig.~\ref{fig:qanloc}. It can be found that the performance of QAN improves at the beginning and reaches the top accuracy at \texttt{Pool3}. The end-to-end training version of feature generation part with quality generation part performs better than that of fixed. So we can make the conclusion that 1) the middle level feature is better for QAN to learn and 2) significant improvement can be achieved by jointly training feature generation part and quality generation part.
\section{Conclusion and future work}
\label{diss}
In this paper we propose a Quality Aware Network (QAN) for set-to-set recognition. It automatically learns the concept of quality for each sample in a set without supervised signal and aggregates the most discriminative samples to generate set representation. We theoretically and experimentally demonstrate that the quality predicted by network is beneficial to set representation and better than human labelled. 

QAN can be seen as an attention model that pay attention to high quality elements in a image set. However, an image with poor quality may still has some discriminative regions. Considering this, our future work will explore a fine-grained quality aware network that pay attention to high quality regions instead of high quality images in a image set.

% ref
\newpage
\begin{spacing}{0.1}

\footnotesize
\bibliography{mybib}
\bibliographystyle{plain}

%-------------------------------------------------------------------------

\end{spacing}
\end{document}